\documentclass[10pt,twocolumn,letterpaper]{article}

\usepackage{wacv}
\usepackage{times}
\usepackage{epsfig}
\usepackage{graphicx}
\usepackage{amsmath}
\usepackage{amssymb}
\usepackage{booktabs}
\usepackage{cuted}
\usepackage{flushend}
\usepackage{comment}
\usepackage{colortbl}
\usepackage{array,multirow}
\usepackage[accsupp]{axessibility}
\usepackage{authblk}
\newcolumntype{P}[1]{>{\centering\arraybackslash}p{#1}}
\newcommand*{\affaddr}[1]{#1} 
\newcommand*{\affmark}[1][*]{\textsuperscript{#1}}
\newcommand*{\email}[1]{\texttt{#1}}
\def\wacvPaperID{1803} 

\wacvfinalcopy 

\ifwacvfinal
\usepackage[breaklinks=true,bookmarks=false]{hyperref}
\else
\usepackage[pagebackref=true,breaklinks=true,colorlinks,bookmarks=false]{hyperref}
\fi
\usepackage{todonotes}
\usepackage{subcaption}
\usepackage[export]{adjustbox}

\begin{document}

\title{
Rebalancing gradient to improve 
self-supervised
co-training of depth,
odometry
and optical flow predictions}

\author{%
Marwane Hariat\affmark[1], Antoine Manzanera\affmark[1], David Filliat \affmark[1, 2]\\
\affaddr{\affmark[1]U2IS, ENSTA Paris, Institut Polytechnique de Paris, Palaiseau, France}\\
\affaddr{\affmark[2]INRIA FLOWERS}\\
\email{\{marwane.hariat, antoine.manzanera, david.filliat\}@ensta-paris.fr}\\
}
\maketitle
\thispagestyle{empty}

\begin{abstract}
We present \textbf{CoopNet}, an approach that improves the cooperation of co-trained networks by dynamically adapting the 
apportionment
of gradient, to ensure equitable learning progress.
It is applied to motion-aware self-supervised prediction of depth maps, by introducing a new hybrid loss, based on a distribution model of photo-metric reconstruction errors made by, on the one hand the depth + odometry paired networks, and on the other hand the optical flow network. This model essentially assumes that the pixels from moving objects (that must be discarded for training depth and odometry), correspond to those where the two reconstructions strongly disagree.
We justify this model by theoretical considerations and experimental evidences. 
A comparative evaluation on KITTI and CityScapes datasets shows that CoopNet improves or is comparable to the state-of-the-art in depth, odometry and optical flow predictions. Our code is available here: \textit{\textbf{https://github.com/mhariat/CoopNet}}.
\end{abstract}


\section{Introduction}

Humans are amazingly competent at inferring 3D structures of a scene from
monocular
images.
This ability is acquired from the very first day of their lives, when infants learn to understand the geometric properties of their environment and  its regularities. Then, they learn to interpret 2D images as 3D scenes by making their visual perception consistent with their inner understanding of the world. 

This mechanism can be emulated with 
self-supervised
learning. To do so, intermediate visual tasks such as depth, optical flow and camera pose estimations are 
performed by deep neural networks
to reproduce a scene from different viewpoints. This whole pipeline can be trained in an end-to-end manner, 
using the consistency between the observed images and synthesised views 
as the supervisory signal \cite{Zhou, Garg}. It will only perform well if the intermediate estimations are close enough to their ground truth.  

Self-supervision has the advantage that the underlying process producing the visual tasks is more robust and can generalise better to new unknown data compared to direct supervision setting with available ground-truth data \cite{Eigen}, where it can be hard to avoid over-fitting due to the lack of constraints. Self-supervised networks have to develop both geometric and contextual reasoning skills, attributes that are far less dataset dependent, to correct the inconsistencies of the view reconstruction. Self-supervision, on top of its healthy training conditions, brings a lot of flexibility. It allows to learn from a much larger scope of data as ground-truth data are not required. Fine-tuning can 
additionally
be stacked 
within
an incremental learning strategy with only minor manageable time and memory increases. 

In our work, we are particularly interested in depth estimation. However, other intermediate visual estimation such as odometry or optical flow will also be considered and assessed with appropriate metrics. We will use only monocular images in order to force visual task estimations to leverage contextual information as much as possible to solve ambiguities, and because it only requires a cheap and ubiquitous monocular camera. 

The view synthesis training strategy requires to face situations such as texture changes, light reflections and occlusions amongst others. But the most challenging issue would certainly be to deal with moving objects,
since the
warping transformation assumes the scene to be static,
and
moving regions can pollute the learning process with misleading high reconstruction errors.

Our contribution is to 
propose a new strategy relying on the cooperation between
the Optical Flow, the Depth and the Pose networks 
during the learning process. 
Basically, regions
for which these networks disagree on 
their view syntheses
are removed from the training samples when necessary. Our completely self-supervised training strategy is assessed 
on 
KITTI \cite{kitti} and Cityscapes \cite{cityscapes}. 
Although simple,
our method outperforms the current state-of-the art unsupervised training strategies dealing with moving objects
by
a substantial margin. It also competes with methods that make use of semantic information coming from off-the-shelf algorithms.

\section{Related Work}

\noindent
\textbf{Self-supervised Learning framework.}
Recently, many research works on unsupervised monocular depth prediction have emerged with the willingness to reduce the gap with fully-supervised methods. The principle is based on the warping image transformation procedure \cite{Garg, Jaderberg}. A target view 
at time
\(t\) is reconstructed from a source view 
at time
\(s\) of the same scene
by calculating a warped image $\hat{I_s}$.
The chosen sources 
timestamps
\(s\) are surrounding the target 
one
\(t\), and generally set to \(\Big\{t - 1, t + 1\Big\}\). 
The
supervision signal used to train the neural network is:



\begin{equation} \label{eq:loss}
\mathcal{L} = \sum_{\sigma} \sum_s \sum_{p_{\sigma}} \Phi \Big(I^{\sigma}_{t}(p_{\sigma}), \hat{I^{\sigma}_{s}}(p_{\sigma}) \Big)
\end{equation}

\noindent
With the \textit{photo-metric 
error
function} {\boldmath $\Phi$} defined as:

\begin{equation} \label{eq:phi}
\Phi \left(x, y\right) = \alpha\frac{1 - SSIM\left(x, y\right)}{2} + \left(1 - \alpha\right)\lvert x - y \rvert
\end{equation}

\noindent
where \textit{SSIM} is the structural-similarity \cite{ssim} and $\sigma$ is a scale index,
since
intermediate downscale estimations are also considered in the process 
to address
the \textit{gradient locality problem} caused by the bilinear interpolation \cite{Jaderberg}. Here 
$I^{\sigma}$
refers to the resized version of image 
$I$
with a downscale factor of 
$\frac{1}{2^{\sigma}}$, and $p_{\sigma}$
is the pixel index of images resized at scale 
$\sigma$.
In the 
remainder
of the paper, we will drop the 
$\sigma$
for better readability.

Now, depending on the objective, the
warped image 
$\hat{I_s}$
can be obtained in two different ways. One introduced by \cite{Zhou} and using the combination of a depth network \(\mathcal{D}_{\theta}\) and a camera pose network \(\mathcal{T}_{\alpha}\), to apply the re-projection formula:


\begin{equation} \label{eq:warp_depth}
\begin{gathered}
\hat{I}^{\theta,\alpha}_s(p) = I_{s} \left( K \hat{T}_{t \rightarrow s} \hat{D}_t(p) K^{-1} p \right)\\
\hat{T}_{t \rightarrow s} = \Big[\mathcal{R}, t\Big] \in \mathcal{SE}\left(3\right)
\end{gathered}    
\end{equation}

\noindent
where $K$ is the calibration matrix, $\hat{T}$ is the displacement matrix predicted by $\mathcal{T}_{\alpha}$, and $\hat{D}$ is the depth map predicted by $\mathcal{D}_{\theta}$.

\noindent
And another one \cite{Meister, Yu} using an \textit{optical flow} network \(\mathcal{F}_{\delta}\) that directly predicts the displacement
vector $F_{\delta}$:



\begin{equation} \label{eq:warp_flow}
\hat{I}^{\delta}_s(p) = I_{s} \left( p + F_{\delta}(p) \right)
\end{equation}



\noindent
\textbf{Accounting for Motion.} 
Unlike the warping of Eq.~\ref{eq:warp_flow}, which does not care for the origin of motion, the warping of Eq.~\ref{eq:warp_depth}
is no longer valid in moving regions, corresponding to objects that have a displacement on their own. 
It then makes sense to define the {\em rigid flow} as the apparent motion flow induced by the camera motion only, under rigid assumption, and calculated as:  
\begin{equation} \label{eq:rigid_flow}
F_{\theta,\alpha}(p) = K \hat{T}_{t \rightarrow s} \hat{D}_t(p) K^{-1} p - p
\end{equation}

Inside moving object regions,
even though depth and pose predictions are correct, 
the \textit{photo-metric Loss} {\boldmath $\Phi$} will 
render wrong 
values and disrupt the back-propagation process
within pose and depth networks. 
There are two ways to fix this issue. Either one adds a residual correction to \(\hat{T}_{t \rightarrow s}\) in order to account for potentially moving objects as done by~\cite{Li} and \cite{Gordon}. Or one can also decide to remove moving object pixels from the loss \(\mathcal{L}\) in Eq.~\ref{eq:loss}. This is the strategy that we decide to follow in this paper.





Being able to detect the moving regions of an image is a real challenge. Hence, several methods \cite{Casser, Lee2} chose to rely on an off-the-shelf instance segmentation algorithms \cite{mask_rcnn} to get rid of potential moving objects. A strong limitations here is the lack of generalisation. Indeed, these off-the-shelf algorithms are trained on different datasets \cite{coco}, as the mainstream ones used in monocular depth estimation don't offer enough annotated ground-truth data. Some works tried to overcome this issue either by incorporating the instance segmentation part into the learning pipeline \cite{Klingner}, with the off-the-shelf algorithm predictions used as the ground-truth data. Or by using the feature maps of the off-the-shelf network to drive \cite{Li3, Guizilini2} the different visual task networks. In both cases the issue still remains.
Besides, these methods are not compliant with our
fully
self-supervised learning setting. We want to be able to keep learning on the fly and benefit from a large scope of data.

Closely related to our work, \cite{Chen} warps an image using the two orthogonal ways from Eq.~\ref{eq:warp_depth} and Eq.~\ref{eq:warp_flow}. The supervision signal is then modified as:


\begin{equation} \label{eq:glnet}
\mathcal{L}_{GLNet} = \sum_{s, p} \Psi \left(I_{t}\left(p\right), \hat{I}^{\theta,\alpha}_{s}\left(p\right), \hat{I}^{\delta}_{s}\left(p\right)\right)
\end{equation}
with $\hat{I}^{\theta,\alpha}_{s}$ and $\hat{I}^{\delta}_{s}$ respectively given by Eq.~\ref{eq:warp_depth} and Eq~\ref{eq:warp_flow}, and \(\Psi\) is the \textit{adaptive photo-metric loss}:

\begin{equation} \label{eq:adaptive_loss}
\Psi\left(x, y, z\right) = \min\left(\Phi\left(x, y\right), \Phi\left(x, z\right)\right)
\end{equation}



This approach therefore tries to detect moving pixels by the difference between the optical flow and rigid flow predictions, assuming a worse prediction by the rigid flow. Other approaches such as \cite{Yang, Yin, Liu, Li2, Lee, Gao} propose to infer a moving object mask using a pre-determined metric related to the geometric inconsistency between the optical flow and the rigid flow. 

Following the idea of~\cite{Chen}, our contribution, rather, incorporates a loss-oriented component as part of the decision on moving pixels, while taking care of the different progression speeds between networks to make them benefit from each other in the best way. Additionally, we continuously adapt our decision criterion along the training process using a quantile based approach.

\section{Limits of the adaptive photo-metric loss}
\subsection{Instability} \label{subsection:instability}

The goal of the \textit{adaptive photo-metric loss} 
of \cite{Chen}
is to co-train,
on the one hand
the pair \(\left(\mathcal{D}_{\theta}, \mathcal{T}_{\alpha}\right)\) 
and, 
on the other hand
\(\mathcal{F}_{\delta}\).
Since the loss distributes the pixel errors to both networks, according to $\arg \min_{y,z} \left( \Phi(x,y),\Phi(x,z) \right)$, the networks are actually competing against each other.
All things being equal, the optical flow \(\mathcal{F}_{\delta}\) is intrinsically better at learning from the photo-metric loss than \(\left(\mathcal{D}_{\theta}, \mathcal{T}_{\alpha}\right)\), as the re-projection (Eq.~\ref{eq:warp_depth}) is more constrained compared to Eq.~\ref{eq:warp_flow}. We will call this property the \textit{intrinsic bias} throughout the paper. This unbalanced learning capacity between the two contestants is worsen over the training epochs. Indeed, the {\boldmath $\Psi$} operator splits the set of pixels in two parts based on the sign of the random variable {\boldmath $\Delta$} defined as:

\begin{equation} \label{eq:delta}
\Delta\left(p\right) = \Phi \Big(I_{t}(p), \hat{I}^{\theta,\alpha}_{s}\left(p\right) \Big) - \Phi \Big(I_{t}(p), \hat{I}^{\delta}_{s}\left(p\right) \Big)
\end{equation}

The probability density function \(f_{\Delta}\) is approximately Gaussian. In \cite{Chen}, pixels for which {\boldmath $\Delta$} has non-zero negative values are used to train the pair \(\left(\mathcal{D}_{\theta}, \mathcal{T}_{\alpha}\right)\), whereas the rest of the pixels train \(\mathcal{F}_{\delta}\). 

\begin{figure}[h]
\centering
  \scalebox{0.19}{\includegraphics{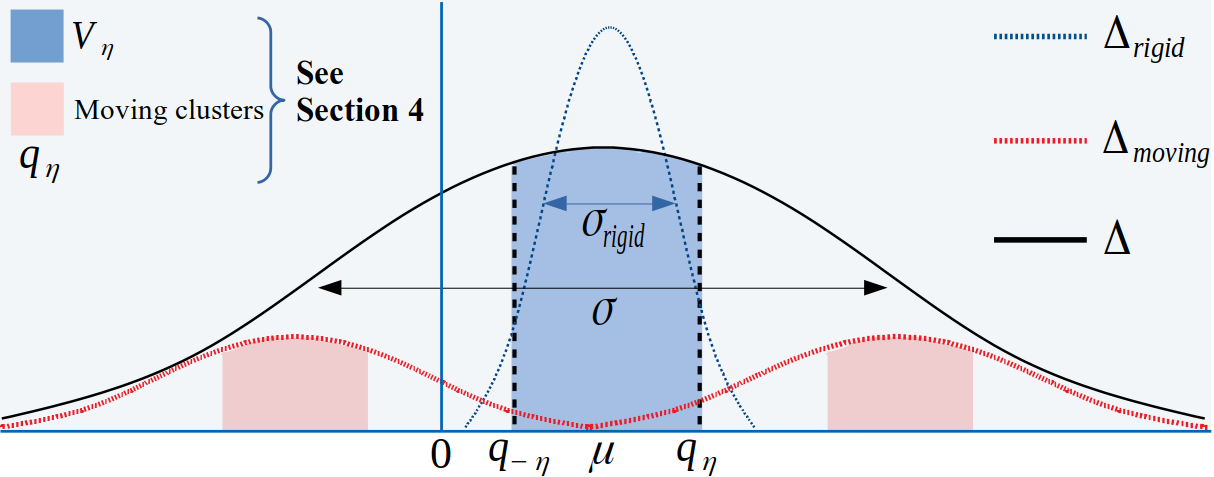}}
  \caption{
  Density models of $\Delta$ used in our work, for all the pixels (black), rigid pixels (blue dashed), and mobile pixels (red dashed). This is the result of the statistical analysis of $\Delta$ on all the images of KITTI and highlights the intrinsic bias of the Gaussian distribution, the moving pixels following a bimodal distribution centred on both sides of the tails and the rigid pixels located around the mean value.
  Note that since rigid pixels are the vast majority, $\mu_{\text{rigid}} \approxeq \mu$.
  }
  \label{fig:sigma_illustration}
\end{figure}

\begin{figure}[h]
\centering
  \scalebox{0.38}{\includegraphics{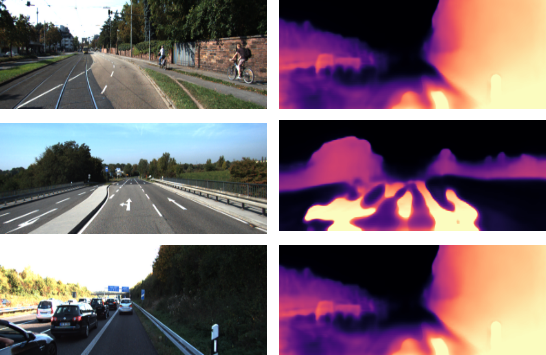}}
  \caption{Degenerate cases with black 
  stains (corresponding to infinite depths)
  spreading all over the image. 
  }
  \label{fig:degenerate}
\end{figure}

Over the training iterations,
\(\mathcal{F}_{\delta}\) takes advantage of its better learning abilities over \(\left(\mathcal{D}_{\theta}, \mathcal{T}_{\alpha}\right)\), 
shifting \(f_{\Delta}\) to the right as shown in Fig.~\ref{fig:sigma_illustration}, 
thus
creating an imbalance on the number of pixels allocated to
each contestant.
It benefits the optical flow network, which 
gets even better at the expense of the depth and pose networks. The resulting sequence of mean values \(\left(\mu_{n} = \mathbb{E}\left[\Delta_{n}\right]\right)_{n \in \mathbb{N}}\), where \(n\) refers to the training iteration index, has an upward trend that needs to be kept under control
to avoid a degenerative state where the optical flow
is
too good and prevents the pair \(\left(\mathcal{D}_{\theta}, \mathcal{T}_{\alpha}\right)\) from learning anything.
The criterion used to study stability is the convergence of the sequence \(\left(\theta_{n}= 1/\mathbb{P}\big(\Delta_{n} < 0\big)\right)_{n \in \mathbb{N}}\), with $\mathbb{P}$ the probability measure. We
provide in supplementary material a
proof that the operator {\boldmath $\Psi$} can make \(\theta_{n}\) diverge if the intrinsic power of \(\mathcal{F}_{\delta}\) is not taken care of.
Practically, the procedure is very sensitive to small changes, especially when it advantages the optical flow. For the same hyper-parameter settings, the depth network can, depending on the initialized weights, either give good predictions or produce bad map estimations as displayed in Fig.~\ref{fig:degenerate}. 

\begin{figure*}[h]
\centering
  \scalebox{0.5}[0.4]{\includegraphics{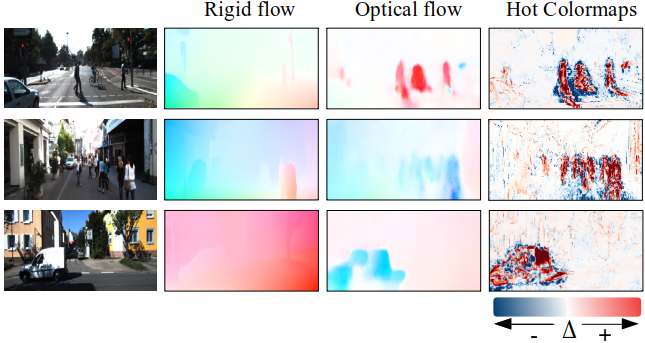}}
  \caption{Smoothing issue around moving objects. 
  The
  \textit{Hot Colormaps} images
  aim at representing both the sign and absolute value of the ${\boldmath \Delta}$ function 
  (see colour bar).}
  \label{fig:mov_issues}
\end{figure*}

\subsection{Fundamental issue}

As illustrated in Fig.~\ref{fig:sigma_illustration}, values of moving pixels {\boldmath $\Delta_{\text{moving}}$} are particularly found in the tails of the distribution. Values in the right part of the tail are due to the systemic inability of the pair \(\left(\mathcal{D}_{\theta}, \mathcal{T}_{\alpha}\right)\) to account for any moving displacement. Values in the left part of the tail often corresponds to moving objects for which the optical flow faces smoothing issues as pictured in Fig. \ref{fig:mov_issues}, rather than a much better prediction of the depth and pose networks. Together, they are responsible for a great part of the variation of {$\Delta$} and thus lead to \(\sigma_{\text{rigid}} < \sigma \). Values of rigid pixels {\boldmath $\Delta_{\text{rigid}}$}, rather, are mostly located in a close neighbourhood of \(\mu\). Intuitively, both the pair \(\left(\mathcal{D}_{\theta}, \mathcal{T}_{\alpha}\right)\) and \(\mathcal{F}_{\delta}\) \textbf{\textit{have a consistent understanding}} for static regions of a scene. Hence, the resulting values taken by {\boldmath $\Delta$} are more stationary and fairly close, neglecting the intrinsic bias. \\
As mentioned previously, the idea of ~\cite{Chen} is to consider pixels \(p\) for which \(\Delta\left(p\right)\) has 
negative values to train the pair \(\left(\mathcal{D}_{\theta}, \mathcal{T}_{\alpha}\right)\). However, by doing so, not only does it throw away a substantial number of rigid pixels, as the  
interval \(\left[\mu - \sigma_{\text{rigid}}, \mu + \sigma_{\text{rigid}}\right]\),
which covers the values of about 68\% of the rigid pixels, satisfies:

\begin{equation} \label{eq:cond_theta}
\begin{aligned}
]-\infty,0[
\cap \left[\mu, \mu + \sigma_{\text{rigid}} \right] 
= \emptyset \\
]-\infty,0[ \cap \left[\mu - \sigma_{\text{rigid}}, \mu \right] 
= \emptyset \text{ if }
\sigma_{\text{rigid}} < \mu
\end{aligned}
\end{equation}

\noindent
but more importantly, it  wrongly takes into account the moving pixels in the {\em left} part of the tails (see Fig.~\ref{fig:sigma_illustration}).

\section{Method}

\begin{figure*}[h]
  \centering
  \scalebox{0.5}[0.49]{\includegraphics{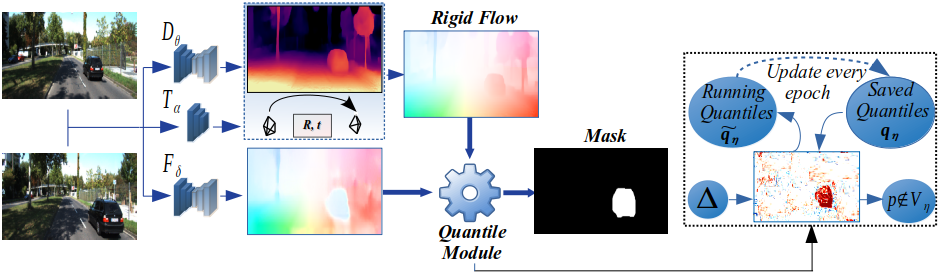}}\quad
  \caption{Diagram depicting  \textbf{CoopNet}. The \textbf{\textit{Quantile Module}} takes as input the rigid flow inferred by the pair \(\left(\mathcal{D}_{\theta}, \mathcal{T}_{\alpha}\right)\) and the flow produced by \(\mathcal{F}_{\delta}\) to compute \boldmath $\Delta$. The running values \(\left(\widetilde{q_{-\eta}}, \widetilde{q_{\eta}}\right)\) are updated with the \(P^2\) algorithm \cite{Quantile} to be used at the next epoch. The current values \(\left(q_{-\eta}, q_{\eta}\right)\) determine the neighbourhood \(\mathcal{V_{\eta}}\) to induce a mask map \{\(p \notin \mathcal{V}_{\eta}\)\}}.
  \label{fig:method}
\end{figure*}

The main purpose of our work is to offer a healthier learning protocol to co-train \(\left(\mathcal{D}_{\theta}, \mathcal{T}_{\alpha}\right)\) and \(\mathcal{F}_{\delta}\). Inspired by the different 
issues raised
in the previous section we propose a \textbf{\textit{Quantile Based Split}} of the probability density function \(f_{\Delta}\) 
in order to:
\begin{itemize}
  \item Only train the pair \(\left(\mathcal{D}_{\theta}, \mathcal{T}_{\alpha}\right)\) on a tight neighbourhood of \(\mu\), and stop considering pixels with values in the left tail, thus focusing better on static pixels.
  \item Train the flow on the whole set of pixels with a larger weight on pixels with {\boldmath $\Delta$} values in the tails of the distribution.
\end{itemize}

The diagram of our method, named \textbf{CoopNet}, is given in Fig.~\ref{fig:method}. Unlike~\cite{Chen}, which makes the networks compete against one another, our approach is based on \textbf{\textit{cooperation}}. In the same spirit as \textbf{\textit{teacher - student techniques}}, the pair \(\left(\mathcal{D}_{\theta}, \mathcal{T}_{\alpha}\right)\) wait for the approval of the stronger network \(\mathcal{F}_{\delta}\) to select pixels to be trained on. If the optical flow \(\mathcal{F}_{\delta}\) agrees with the pair \(\left(\mathcal{D}_{\theta}, \mathcal{T}_{\alpha}\right)\) on the displacement of a given pixel \(p\), then this pixel \(p\) can safely be considered as rigid and used to feed the loss \(\mathcal{L}_{\theta, \alpha}\) defined in Eq. \ref{eq:loss_cooptnet} below.

\subsection{Description}

For $\eta \in [0,0.5]$, let us denote
\(q_{\eta}\) the 
$(0.5 + \eta)$-quantile
of the probability density function \(f_{\Delta}\). 
Let us also
define \(\mathcal{V}_{\eta} = \left[q_{-\eta}, q_{\eta}\right]\) a neighbourhood of \(\mu\). 

As said in the previous section, most of the rigid pixels 
have a ${\boldmath \Delta}$ value close to
\(\mu\). That's why in our approach, the pair \(\left(\mathcal{D}_{\theta}, \mathcal{T}_{\alpha}\right)\) is only trained on pixels 
belonging to ${\boldmath \Delta}^{-1}\left(\mathcal{V}_{\eta}\right)$
(see Fig.~\ref{fig:sigma_illustration}), with $\Delta^{-1}$ the inverse image.
The larger the interval \(\mathcal{V}_{\eta}\), the closer to the tails and the more likely 
the pollution by
large absolute values {\boldmath $\lvert\Delta\left(p\right)\rvert$} of moving pixels, which is not desirable. On the contrary, a small \(\mathcal{V}_{\eta}\) will filter many pixels and the pair \(\left(\mathcal{D}_{\theta}, \mathcal{T}_{\alpha}\right)\) may not learn anything, as the back-propagation needs enough samples to work. Hence, the hyper-parameter \(\eta\) has to be adjusted to find the best trade-off. 

Different training strategies were experimented to find the best way to train the optical flow network,
and
we found out 
that
learning from all 
the pixels 
was
the most effective manner, with a 
weighted sum advantaging moving pixels. Pixels \(p\) corresponding to values \(\Delta\left(p\right)\) in the tails, specifically the \(\eta\)-quantile and the \(\left(1 - \eta\right)\)-quantile have more weights (Eq.~\ref{eq:w_lossdelta}). 

These two ideas lead to a split of the loss \(\mathcal{L}\) (eq~\ref{eq:loss}), 
into
two terms, as follows:

\begin{equation} \label{eq:loss_cooptnet}
\begin{gathered}
\mathcal{L}_{\delta} = \sum_{p \in \mathcal{P}} w\left(p\right)\Phi \Big(I_{t}(p), \hat{I}^{\delta}_{s}\left(p\right) \Big)\\
\mathcal{L}_{\theta, \alpha} = \sum_{p \in \boldmath{\Delta}^{-1}\left(\mathcal{V}_{\eta}\right)} \Phi \Big(I_{t}(p), \hat{I}^{\theta,\alpha}_{s}\left(p\right) \Big)\\
\mathcal{L}_{\text{CoopNet}} = \mathcal{L}_{\theta, \alpha} + \mathcal{L}_{\delta}
\end{gathered}    
\end{equation}

\noindent
with $\mathcal{P}$ the set of all pixels and \(w\) defined as:

\begin{equation} \label{eq:w_lossdelta}
\begin{gathered}
w\left(p\right)= \begin{cases} \frac{\lvert\mathcal{P}\rvert}{\lvert\boldmath{\Delta}^{-1}\left(\Gamma\right)\rvert} & \text{if}\ p \in \boldmath{\Delta}^{-1}\left(\Gamma\right)\\
\frac{\lvert\mathcal{P}\rvert}{\lvert\boldmath{\Delta}^{-1}\left(\overline{\Gamma}\right)\rvert} &  \text{otherwise}
\end{cases}\\
\Gamma=]-\infty,q_{-\eta}[ \cup ]q_{\eta},+\infty[
\end{gathered}
\end{equation}

\noindent
The losses \(\mathcal{L}_{\theta, \alpha}\) and \(\mathcal{L}_{\delta}\) are used to train respectively the pair \(\left(\mathcal{D}_{\theta}, \mathcal{T}_{\alpha}\right)\) and \(\mathcal{F}_{\delta}\).

\subsection{Advantages over the adaptive photo-metric loss}

Training the optical flow on a different set, the complement \(\overline{\mathcal{V}_{\eta}}\) of the one used by the pair \(\left(\mathcal{D}_{\theta}, \mathcal{T}_{\alpha}\right)\) for instance, as done by \cite{Chen}, would be sub-optimal. The performances of the optical flow network on rigid regions would be very poor, close to a random prediction. As a consequence, rigid pixels \(p_{\text{rigid}}\) would have significant negative {\boldmath $\Delta\left(p_{\text{rigid}}\right)$} values and would be mixed with the large negative values of moving pixels that \(\mathcal{F}_{\delta}\) failed to predict correctly. Although the most important is to distinguish rigid pixels from moving pixels correctly predicted by \(\mathcal{F}_{\delta}\) which are in much greater numbers, this is not ideal. With our approach, rather, the probability density function \(f_{\Delta}\) has a clumped dispersion pattern as illustrated in Fig. \ref{fig:sigma_illustration}, with three clusters sharply delimited. And the intersection between the rigid cluster and each of the two other moving ones is greatly limited. Keeping all that has been said so far in mind, one can legitimately wonder how the distribution of {\boldmath $\Delta$} can stay clamped with three clusters in the approach of \cite{Chen}, as stated in the previous section. The optical flow is trained on 
$\boldmath{\Delta}^{-1} \left( ]-\infty,0[ \right)$
a subset which is, luckily, composed of enough rigid pixels for \(\mathcal{F}_{\delta}\) to be decent on the static regions. Unfortunately this property is not taken advantage of thereafter. 

Finally, let us define:
\begin{equation} \label{eq:expect}
\begin{gathered}
L_{1} = \mathbb{E}\bigg[\Phi \Big(I_{t}(p), \hat{I}^{\theta,\alpha}_{s}\left(p\right) \Big) \Big|p \in \mathcal{P} \text{,  }\Delta\left(p\right) \in \mathcal{V}_{\eta}\bigg]\\
L_{2} = \mathbb{E}\bigg[\Phi \Big(I_{t}(p), \hat{I}^{\theta,\alpha}_{s}\left(p\right) \Big) \Big|p \in \mathcal{P} \text{,  }\Delta\left(p\right) < 0\bigg]
\end{gathered}
\end{equation}

We give in 
supplementary material
the mathematical proof that \( L_{1} < L_{2}\).
This inequality demonstrates theoretically the benefits of introducing the neighbourhood \(\mathcal{V}_{\eta}\) over 
using
the 
sign of $\boldmath{\Delta}(p)$ like
\cite{Chen}.

\subsection{Regularisation set}
We observe experimentally that too many moving pixels can still pollute the neighbourhood \(\mathcal{V}_{\eta}\), even when the hyper-parameter \(\eta\) is set to be very selective. This is due to the well known weakness of the \textit{photo-metric function}~\(\Phi\)~(Eq.~\ref{eq:warp_flow}): Because \(\Phi\) compares images based on colour similarities, it has difficulties to be discriminative in homogeneous areas. 
Hence, although the pair \(\left(\mathcal{D}_{\theta}, \mathcal{T}_{\alpha}\right)\) and \(\mathcal{F}_{\delta}\) disagree on the displacement to be made to warp a moving pixel \(p\), the value \(\Delta\left(p\right)\) might still be similar to the ones taken by rigid pixels and thus fall 
into
\(\mathcal{V}_{\eta}\). For this reason, we propose to add a new constraint on the agreement of both networks on the pixels displacement.

We introduce \(\Delta_{\text{flow}}\) defined as:

\begin{equation} \label{eq:delta_flow}
\begin{gathered}
\Delta_{\text{flow}}\left(p\right) = \frac{F_{\theta, \alpha}\left(p\right) - F_{\delta}\left(p\right)}{\big\|F_{\theta, \alpha}\left(p\right)\big\|_{2} + \big\|F_{\delta}\left(p\right)\big\|_{2}}
\end{gathered}
\end{equation}

where \(F_{\theta, \alpha}\left(p\right)\) is the image displacement of a pixel produced by the image warping. As for \(\Delta\), the closer the pair \(\left(\mathcal{D}_{\theta}, \mathcal{T}_{\alpha}\right)\) and \(\mathcal{F}_{\delta}\), the 
smaller
\(\Delta_{\text{flow}}\). 
However, the
flow values \(F_{\theta, \alpha}\left(p\right)\) and  \(F_{\delta}\left(p\right)\) are vectors with two components, 
then
\(\Delta_{\text{flow}}\) is a 
2d random vector.
As the flow value has a strong dependency on the position in the image (close pixels tend to have higher flow magnitudes than far pixels), a normalisation term is added in the denominator in Eq.~\ref{eq:delta_flow}. The random variable \(\Delta_{\text{flow}}\) doesn't take into account colour intensities and is thus insensitive to the homogeneous issue raised previously. Besides, it has no intrinsic bias which exists with \(\Delta\) because of the operator \(\Phi\). 

Finally, the neighbourhood \(\mathcal{V}_{\eta}\) chosen to compute \(\mathcal{L}_{\theta, \alpha}\) in Eq.~\ref{eq:loss_cooptnet} can be replaced by \(\mathcal{V}\):

\begin{equation} \label{eq:16}
\begin{gathered}
\mathcal{V} = \mathcal{V}_{\eta} \cap \mathcal{V}_{\zeta}\\
\mathcal{V}_{\zeta} = \mathcal{V}_{\zeta}^{\text{flow, x}} \cap \mathcal{V}_{\zeta}^{\text{flow, y}}
\end{gathered}
\end{equation}

\noindent
with \(\mathcal{V}_{\zeta}^{\text{flow, x}}\) and \(\mathcal{V}_{\zeta}^{\text{flow, y}}\) defined in the same way as \(\mathcal{V}_{\eta}\) using respectively \(\Delta_{\text{flow}}^{x}\) and \(\Delta_{\text{flow}}^{y}\). The loss-oriented neighbourhood \(\mathcal{V}_{\eta}\) remains the main actor of the decision process,
while \(\mathcal{V}_{\zeta}\) can be seen as a \textbf{prior with a regularisation} effect to solve the homogeneous issue. One might well ask why not consider the magnitude of the flow differences instead.
The reason is that this choice makes all that had been demonstrated with \(\Delta\) (that assumes signed values and a normally distributed random variable) to remain true.

\subsection{Implementation details}

\noindent
\textbf{Additional Losses} On top of \(\mathcal{L}_{\text{CoopNet}}\) defined in Eq.~\ref{eq:loss_cooptnet}, we also take into account the following subsidiary losses:

\begin{itemize}
  \item The geometric consistency loss \(\mathcal{L}_{\text{gc}}\) proposed by \cite{Bian}.
  \item A forward-backward consistency check \(\mathcal{L}_{\text{fwd,bwd}}\) of the optical flow \(\mathcal{F}_{\delta}\) as done in \cite{Meister, Zou}.
  \item The standard edge-aware smoothness loss \(\mathcal{L}_{s}\) for both depth and flow maps. The normalised disparity is used here as proposed by \cite{Wang2} to avoid divergence.
  \item The epipolar constraint \(\mathcal{L}_{\text{ep}}\) with different version of the one proposed in \cite{Chen} (see supplementary materials).
  \item The inverse of the variance of depth maps \(\mathcal{L}_{\text{var}}\) mentioned in \cite{Kim} in order to stabilise the training process. 
\end{itemize}

The final loss is:
\begin{equation} \label{eq:19}
\begin{gathered}
\mathcal{L}_{\text{final}} = \mathcal{L}_{\text{CoopNet}} + \lambda_{gc} \mathcal{L}_{gc} + \lambda_{\text{fwd,bwd}} \mathcal{L}_{\text{fwd, bwd}}\\
+ \lambda_{s} \mathcal{L}_{s} + \lambda_{\text{ep}} \mathcal{L}_{\text{ep}} + \lambda_{\text{var}} \mathcal{L}_{\text{var}}
\end{gathered}
\end{equation}

\noindent
\textbf{Network Architectures.} Our focus in this work is to promote our \textbf{\textit{cooperation}} learning protocol and to see how it compares with the other well-established \textbf{self-supervised depth estimation} training strategies \cite{Godard, Zhou, Chen, Li}. For a better comparison, we decided to use the same standard networks as those methods. In particular, for both the depth and flow networks, we adopt a UNet structure with four intermediate multi-scale predictions as proposed by \cite{Zhou}. The Pose networks is based on a ResNet encoder at the end of which a 6-DoF vectors is predicted. For the depth network we use the specific DispResNet architecture of \cite{Godard}. For the flow network we implement the ResFlowNet of \cite{Yin, Lee}. Both the depth and pose networks have a ResNet18 backbone while the flow network uses a ResNet50 encoder. More efficient networks 
could
of course improve performances even more. For instance, regarding the depth network, PackNet \cite{Guizilini}, architectures using attention \cite{Johnston, Gao, Jia, Li2} and/or cost-volume \cite{Watson, Johnston} are 
performing well, just like 
FlowNet \cite{Fischer} and PwC-Net \cite{Hur, PwC} for the flow prediction.

\noindent
\textbf{Occlusions} are dealt with in two ways. The \textit{warping module} of \cite{Wang} is used to mask occluded pixels in \(\mathcal{L}_{\delta}\) with a hard occlusion threshold set to \(0.2\). While for \(\mathcal{L}_{\theta, \alpha}\), occluded pixels are handled thanks 
to
the standard \textit{minimum re-projection} of \cite{Godard}.


\begin{table*}[ht]
\centering
\begin{tabular}{ |P{0.4cm}|p{2.3cm}|c|c|c|c|c|c|c|c|}
\hline
Set 
&Method&Size&\cellcolor{red!30}Abs Rel&\cellcolor{red!30}Sq Rel&\cellcolor{red!30}RMSE&\cellcolor{red!30}RMSE log&\cellcolor{blue!30}\(\delta < 1.25\)&\cellcolor{blue!30}\(\delta < 1.25^2\)&\cellcolor{blue!30}\(\delta < 1.25^3\)\\
\hline
\centering \parbox[t]{5mm}{\multirow{9}{*}{{

\textbf{K}}}}&Li \textit{et al.} \cite{Li} &\(128 \times 416\)&0.130&\textbf{0.950}&5.138&0.209&0.843&0.948&0.978\\
&DLNet\cite{Jia} \(\ddagger\) &\(128 \times 416\)&0.128&0.979&\textbf{5.033}&\underline{0.202}&.851&0.954&0.980\\
&\textbf{CoopNet}&\(128 \times 416\)&\underline{0.126}&1.014&5.091&0.204&\underline{0.856}&\underline{0.954}&\textbf{0.980}\\
&\textbf{CoopNet R50}&\(128 \times 416\)&\textbf{0.121}&\underline{0.971}&\underline{5.055}&\textbf{0.199}&\textbf{0.863}&\textbf{0.955}&\textbf{0.980}\\
\cline{2-10}
&Monodepth2\cite{Godard}&\(192 \times 640\)&\underline{0.115}&0.903&4.863&0.193&0.877&0.959&0.981\\
&SGDepth \cite{Klingner} \(\dagger\)&\(192 \times 640\)&0.117&0.907&\textbf{4.693}&\underline{0.191}&\textbf{0.879}&\textbf{0.961}&0.981\\
&Tosi \textit{et al.} \cite{Tosi} \(\dagger\) &\(192 \times 640\)&0.126&\textbf{0.835}&4.937&0.199&0.844&0.953&0.982\\
&\textbf{CoopNet}&\(192 \times 640\)&\textbf{0.113}&\underline{0.872}&\underline{4.824}&\textbf{0.190}&\underline{0.878}&\underline{0.959}&\textbf{0.982}\\
\cline{2-10}
&Insta-DM \cite{Lee2} \(\dagger\) &\(256 \times 832\)&0.112&0.777&4.772&0.191&0.872&0.959&0.982\\

\hline
\hline
\centering \parbox[t]{10mm}{\multirow{4}{*}{{
\textbf{CS}}}}&Struct2Depth\cite{Casser}\(\dagger\)&\(128 \times 416\)&0.145&1.737&7.28&0.205&0.813&0.942&0.978\\
&Gordon\cite{Gordon} \(\dagger\) &\(128 \times 416\)&0.127&\underline{1.330}&\underline{6.96}&0.195&0.830&0.947&\underline{0.981}\\
&Li \textit{et al.}\cite{Li}&\(128 \times 416\)&\textbf{0.119}&\textbf{1.29}&\textbf{6.98}&\textbf{0.190}&\textbf{0.846}&\textbf{0.952}&\textbf{0.982}\\
&\textbf{CoopNet}&\(128 \times 416\)&\underline{0.121}&1.443&7.01&\textbf{0.190}&\textbf{0.846}&\underline{0.951}&0.980\\
\hline
\end{tabular}
\caption{\textbf{Results of depth estimations}. We only compare our methods to the most recent and competitive algorithms. All results here are presented for different image sizes. For each metric the best result is displayed in bold and the second one is underlined. The depth cutoff is set to 80m. For red metrics, lower is better. For blue metrics, higher is better.
\(\boldsymbol{\dagger}\) - Use of an off-the-shelf semantic algorithms.
\(\boldsymbol{\ddagger}\) - Use of a transformer depth network.
\textbf{K}: trained and evaluated on KITTI. \textbf{CS}: trained and evaluated on Cityscapes. \textbf{R50:} Use a ResNet50 backbone instead of ResNet18 for the depth network.}
\label{table:results_depth}
\end{table*}

\begin{table*}
\begin{minipage}[t]{\columnwidth}
\centering
\begin{tabular}{|c|c|c|c|c|}
\hline
\multirow{2}{*}{Methods} & \multicolumn{2}{c|}{Seq. 09} & \multicolumn{2}{c|}{Seq. 10} \\
\cline{2-5}
& {\(t_{err}\left(\%\right)\)} & {\(r_{err}\left(^{\circ}/100m\right)\)}  & {\(t_{err}\left(\%\right)\)} & {\(r_{err}\left(^{\circ}/100m\right)\)} \\
\cline{1-5}
ORB\cite{slam}&15.30&0.26&3.68&0.48\\
\text{Zhou}\cite{Zhou}&17.84&6.78&37.91&17.78\\
\text{Bian}\cite{Bian}&11.2&3.35&10.1&4.96\\
\textbf{CoopNet}&\textbf{8.42}&\textbf{2.66}&\textbf{7.29}&\textbf{2.14}\\
\hline
\end{tabular}
\caption{\textbf{Odometry}: Average Translation and Rotation errors for sequence 09 and 10 of the KITTI Odometry Dataset.}
\label{table:results_odometry}
\end{minipage}\hfill 
\begin{minipage}[t]{\columnwidth}
\centering
\begin{tabular}{|c|c|c|}
\hline
Method&Noc&All\\
\hline
FlowNetS\cite{Fischer}&8.12&14.19\\
FlowNet2\cite{flownet2}&\underline{4.93}&10.06\\
GeoNet\cite{Yin}&8.05&10.81\\
GLNet\cite{Chen}&\textbf{4.86}&\textbf{8.35}\\
\textbf{CoopNet}&5.10&\underline{9.43}\\
\hline
\end{tabular}
\caption{{\bf Optical Flow:} Average end point error (in pixels) for non occluded (Noc) and for all (All) pixels on the KITTI 2015 flow dataset.}
 \label{table:results_flow}
  \end{minipage}
\end{table*}

\begin{table}[ht]
\centering
\resizebox{\columnwidth}{!}{%
\begin{tabular}{|c|c|c|c|c|c|c|c|c|c|}
\hline
\multirow{2}{*}{\(\mathcal{L}_{\textit{gc}}\)} & \multirow{2}{*}{\(\mathcal{L}_{\textit{ep}}\)} & \multirow{2}{*}{\(\mathcal{L}_{\textit{s}}\)} & \multirow{2}{*}{\(\mathcal{L}_{\textit{fwd,bwd}}\)} & \multirow{2}{*}{\(\mathcal{L}_{\textit{var}}\)} & \multicolumn{3}{c|}{\(\mathcal{L}_{\textit{photo}}\)} &
\multirow{2}{*}{Abs Rel} &
\multirow{2}{*}{APE}\\
\cline{6-8}

&&&&& {\(\mathcal{L}_{\textit{baseline}}\)} & {\(\mathcal{L}_{\textit{apc}}\)}  & {\(\mathcal{L}_{\textit{CoopNet}}\)}&&\\
\hline
\multicolumn{1}{|c}{}&\multicolumn{1}{c}{}&\multicolumn{1}{c}{\checkmark}&\multicolumn{1}{c}{}&\multicolumn{1}{c}{}&\multicolumn{1}{c}{\checkmark}&\multicolumn{1}{c}{}&\multicolumn{1}{c|}{}&\multicolumn{1}{c}{0.157}&\multicolumn{1}{c|}{11.93}\\
\multicolumn{1}{|c}{}&\multicolumn{1}{c}{}&\multicolumn{1}{c}{\checkmark}&\multicolumn{1}{c}{}&\multicolumn{1}{c}{}&\multicolumn{1}{c}{}&\multicolumn{1}{c}{\checkmark}&\multicolumn{1}{c|}{}&\multicolumn{1}{c}{0.144}&\multicolumn{1}{c|}{12.26}\\
\multicolumn{1}{|c}{}&\multicolumn{1}{c}{}&\multicolumn{1}{c}{\checkmark}&\multicolumn{1}{c}{}&\multicolumn{1}{c}{}&\multicolumn{1}{c}{}&\multicolumn{1}{c}{}&\multicolumn{1}{c|}{\checkmark}&\multicolumn{1}{c}{0.130}&\multicolumn{1}{c|}{9.74}\\
\hline
\multicolumn{1}{|c}{}&\multicolumn{1}{c}{}&\multicolumn{1}{c}{\checkmark}&\multicolumn{1}{c}{\checkmark}&\multicolumn{1}{c}{}&\multicolumn{1}{c}{}&\multicolumn{1}{c}{}&\multicolumn{1}{c|}{\checkmark}&\multicolumn{1}{c}{0.130}&\multicolumn{1}{c|}{9.21}\\
\multicolumn{1}{|c}{}&\multicolumn{1}{c}{}&\multicolumn{1}{c}{\checkmark}&\multicolumn{1}{c}{\checkmark}&\multicolumn{1}{c}{\checkmark}&\multicolumn{1}{c}{}&\multicolumn{1}{c}{}&\multicolumn{1}{c|}{\checkmark}&\multicolumn{1}{c}{0.130}&\multicolumn{1}{c|}{9.16}\\
\multicolumn{1}{|c}{\checkmark}&\multicolumn{1}{c}{}&\multicolumn{1}{c}{\checkmark}&\multicolumn{1}{c}{\checkmark}&\multicolumn{1}{c}{}&\multicolumn{1}{c}{}&\multicolumn{1}{c}{}&\multicolumn{1}{c|}{\checkmark}&\multicolumn{1}{c}{0.128}&\multicolumn{1}{c|}{9.27}\\
\multicolumn{1}{|c}{\checkmark}&\multicolumn{1}{c}{\checkmark}&\multicolumn{1}{c}{\checkmark}&\multicolumn{1}{c}{\checkmark}&\multicolumn{1}{c}{\checkmark}&\multicolumn{1}{c}{}&\multicolumn{1}{c}{}&\multicolumn{1}{c|}{\checkmark}&\multicolumn{1}{c}{0.126}&\multicolumn{1}{c|}{9.43}\\
\hline
\multicolumn{1}{|c}{\checkmark}&\multicolumn{1}{c}{\checkmark}&\multicolumn{1}{c}{\checkmark}&\multicolumn{1}{c}{\checkmark}&\multicolumn{1}{c}{}&\multicolumn{1}{c}{}&\multicolumn{1}{c}{\checkmark}&\multicolumn{1}{c|}{}&\multicolumn{1}{c}{0.135}&\multicolumn{1}{c|}{8.35}\\
\hline
\end{tabular}
}

\caption{\textbf{Ablation study on absolute relative error 
(depth)
and average end point error
(flow).
} Resolution size: \(128 \times 416\). In our baseline, both the optical flow \(\mathcal{F}_{\delta}\) and the pair \(\left(\mathcal{D}_{\theta}, \mathcal{T}_{\alpha}\right)\) are trained using the standard \textit{photo-metric} 
losses
\(\mathcal{L}\) (see 
equations \ref{eq:loss} to \ref{eq:warp_flow}) computed on all of the pixels. 
As common practice,
the smoothness loss \(\mathcal{L}_{\textit{s}}\) is 
used in 
all
experiments. 
Last row corresponds to GLNet \cite{Chen}.
}
\label{table:ablation_study}
\end{table}

\noindent
\textbf{Parameter settings.} 
Our method is implemented in PyTorch. Training is done using the optimiser Adam \cite{Adam} with \(\beta_{1} = 0.99\) and \(\beta_{2} = 0.999\). ResNet backbones are initialised using ImageNet \cite{Imagenet} pretrained weights. Networks are trained for \(30\) epochs with a batch size of 4. The initial learning is set to \(10^{-4}\) and decreased to \(10^{-5}\) after 20 epochs. Standard data-augmentation is performed including horizontal flips, random contrast, saturation, hue 
and brightness
jitters. 
A burning step of 5 epochs is used during which the pair \(\left(\mathcal{D}_{\theta}, \mathcal{T}_{\alpha}\right)\) is trained with \cite{Godard}. Quantiles are computed on the fly based on the algorithm of \cite{Quantile}. The neighbourhood \(\mathcal{V}_{\eta}\) is determined using quantile values of the previous epoch (see Fig.~\ref{fig:method}) with \(\eta = 0.15\) and \(\zeta = 0.25\). The loss weights were determined with a grid-search and finally set to \(\lambda_{\text{gc}} = 0.001\), \(\lambda_{\text{fwd, bwd}} = 0.001\), \(\lambda_{s} = 0.01\), \(\lambda_{ep} = 0.001\) and \(\lambda_{\text{var}} = 10^{-6}\). We employ a single NVIDIA GTX 1080 Ti GPU. Training time takes 12 hours. 

\section{Experiments}

We conducted extensive experiments on depth, camera pose and optical flow estimations in order to validate our method. We present 
the results 
obtained on two datasets:\\

\noindent
\textbf{KITTI} \cite{kitti} is the most popular benchmark to evaluate depth and ego-motion estimations. It consists of 
urban, rural and highway images,
captured by driving around the city of \textit{Karlsruhe}. We use the standard evaluation protocol to retrieve the ground truth depth values from the LIDAR sensor data, and we follow the standard data split proposed by~\cite{Eigen} with 22\,600 training pair images and 697 test pair images.\\

\noindent
\textbf{Cityscapes} \cite{cityscapes} is also composed of urban images but it contains a greater variety of situations with images coming from more than 50 European cities and is challenging because it contains more scenes with moving objects. Following the protocol of \cite{Li} training is done on 22\,973 image-pairs obtained by completing the usual 2975 training images with the 19\,998 extra-training images. For evaluation, we use the 1\,525 test images.

\subsection{Depth}

\begin{figure*}[ht]
\centering
  \scalebox{0.6}[0.55]{\includegraphics{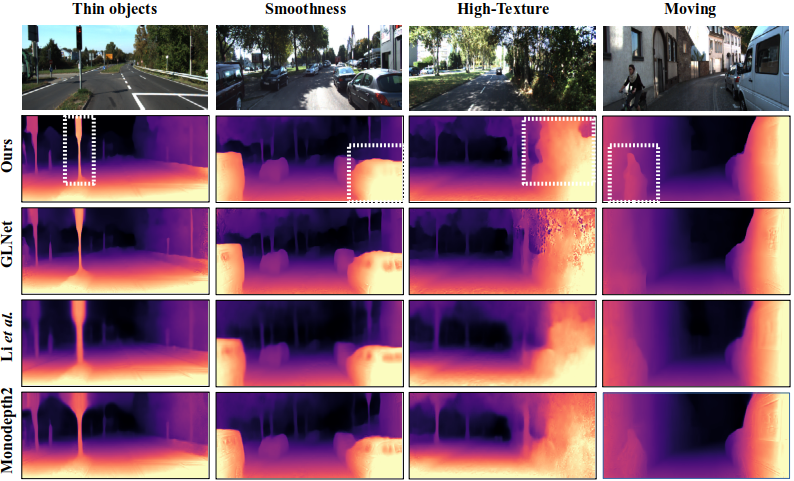}}
  \caption{Comparison of depth map estimation algorithms in challenging situations. 
  White
  Dashed rectangles target the improvement brought by our 
  method.
  }
  \label{fig:depth_compare}
\end{figure*}

Quantitative results are given in Tab.~\ref{table:results_depth}. \textbf{CoopNet} outperforms already very effective methods with a substantial margin in almost all of the different metrics. Qualitative results are presented in Fig.~\ref{fig:depth_compare}. Overall, depth maps yielded by \textbf{CoopNet} are sharpest and achieve better predictions in challenging situation such as thin objects, high texture and moving regions. An ablation study is also presented in Tab.~\ref{table:ablation_study}. Note the great improvements brought by \(\mathcal{L}_{\text{CoopNet}}\) (line 3) compared to the two other types of photo-metric loss (line 1-2). As displayed in the second part of the ablation study, the subsidiary loss benefits are marginal as compared to \(\mathcal{L}_{\text{CoopNet}}\).

\begin{minipage}[c]{0.5\textwidth}
\end{minipage}

\begin{figure*}[h]
\centering
\subfloat[\centering Thin objects]{\scalebox{0.31}[0.45]{\includegraphics{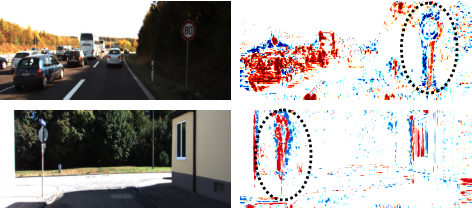}}}
\quad
\subfloat[\centering High-Texture regions]{\scalebox{0.31}[0.45]{\includegraphics{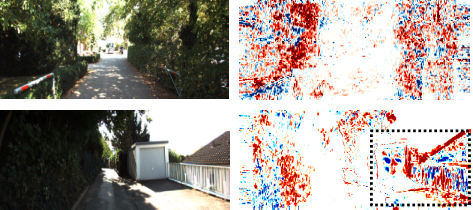}}}
\quad
\subfloat[\centering Edges]{\scalebox{0.31}[0.45]{\includegraphics{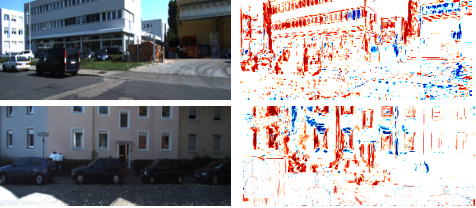}}}
\caption{Illustration of the large variations of {\boldmath $\Delta$} between positive and negative values in 
challenging cases
indicative of a \textbf{strong ambiguity}. Also note the dominance of the red colour due to the intrinsic bias mentioned in Sec.~\ref{subsection:instability}.}
\label{fig:tricky}
\end{figure*}

\subsection{Optical flow and Odometry}

To assess optical flow we use the KITTI 2015 flow dataset containing 200 annotated training images as test images. Tab.~\ref{table:results_flow} shows that \textbf{CoopNet} gives close results to GLNet\cite{Chen} while outperforming all the other methods.\\
The results of our camera-pose estimations trained on KITTI are shown in Tab.~\ref{table:results_odometry}. Again, our method achieves significant gains over the presented methods. We chose to compare specifically to \cite{Bian}, as their \textit{scale-consistent} approach 
is particularly centred on the odometry. The results are however still below classical approaches~\cite{slam}.

\subsection{Visual analysis of $\Delta$}

Fig.~\ref{fig:tricky} shows some example of $\Delta$ values.
We observe experimentally that rigid pixels \(p_{\text{rigid}}\) for which {\boldmath $\Delta\left(p_{\text{rigid}}\right)$} differs the most from \(\mu\) (blue and red values) correspond to tricky cases where it's quite difficult, even for a human, to determine the flow displacement: for instance pixels at the edges, in high-texture areas or around thin objects. 
Conversely, values nearby \(\mu\) (white values) come from rigid pixels that are easy to infer. This supports our claim that when \(\eta\) is low enough, the neighbourhood \(\mathcal{V}_{\eta}\) can be seen as an \textit{agreement area}. In other words, both the pair \(\left(\mathcal{D}_{\theta}, \mathcal{T}_{\alpha}\right)\) and \(\mathcal{F}_{\delta}\) \textit{share the same understanding on the mechanism that governs the displacement} of pixels 
from ${\boldmath \Delta}^{-1} \left(\mathcal{V}_{\eta}\right)$.

\section{Conclusion}
We have presented \textbf{CoopNet}, a training strategy that achieves competitive performances in depth, ego-motion and optical flow estimations using unsupervised training. It relies on a healthy cooperation between different visual tasks so that each one can benefit from the others, relying on the fact that networks should agree on their warping displacement prediction for a pixel to be considered as rigid. 

This idea 
could
be further improved by combining it with an explicit residual correction of the ego-motion \cite{Li}. 
In this case $\mathcal{V}_{\eta}$ no longer represents rigid pixels exclusively, but can still be seen as an \textbf{\textit{agreement area}} from which one can take advantage to emphasise the training process on pixels for which networks disagree.

\newpage

{\small
\bibliographystyle{ieee_fullname}
\bibliography{egbib}
}


\end{document}